\documentclass[10pt,twocolumn,letterpaper]{article}

\usepackage[pagenumbers]{cvpr} 

\usepackage{times}
\usepackage{colortbl}
\usepackage{epsfig}
\usepackage{graphicx}
\usepackage{amsmath}
\usepackage{amssymb}
\usepackage{makecell} 

\usepackage{enumitem} 
\usepackage[linesnumbered,ruled,vlined,algo2e]{algorithm2e}
\usepackage{algorithm, algorithmic}
\usepackage{multirow}
\usepackage{threeparttable}
\usepackage{stfloats}
\usepackage{booktabs}
\usepackage{arydshln}
\usepackage{bm}
\usepackage{makecell}
\usepackage{pifont}
\usepackage{lipsum}
\usepackage{ulem}
\usepackage[table,dvipsnames]{xcolor} 
\usepackage[accsupp]{axessibility} 

\definecolor{cvprblue}{rgb}{0.21,0.49,0.74}
\usepackage[pagebackref,breaklinks,colorlinks,citecolor=cvprblue]{hyperref}

\usepackage[capitalize]{cleveref}
\crefname{section}{Sec.}{Secs.}
\Crefname{section}{Section}{Sections}
\Crefname{table}{Table}{Tables}
\crefname{table}{Tab.}{Tabs.}

\colorlet{colorFst}{Green!25}       
\colorlet{colorSnd}{SpringGreen!45} 
\colorlet{colorTrd}{Yellow!30}      
\colorlet{colorLow}{darkgray!30}    
\newcommand{\fs}{\cellcolor{colorFst}\bf}   
\newcommand{\nd}{\cellcolor{colorSnd}}      
\newcommand{\rd}{\cellcolor{colorTrd}}      
\newcommand{\ours}{\cellcolor[rgb]{.9,.95,.98}} 

\def\eg{\emph{e.g.}\xspace} 
\def\ie{\emph{i.e.}\xspace}

\newcommand{\boldparagraph}[1]{\vspace{0.1em}\noindent{\bf #1}}
\newcommand\blfootnote[1]{%
  \begingroup
  \renewcommand\thefootnote{}\footnote{#1}%
  \addtocounter{footnote}{-1}%
  \endgroup
}

\def\paperID{2976} 
\def\confName{CVPR}
\def\confYear{2024}

\title{MESA: Matching Everything by Segmenting Anything}

\author{Yesheng Zhang \qquad Xu Zhao$^{*}$\\
Department of Automation, Shanghai Jiao Tong University\\
}

\begin{document}
\maketitle
\begin{abstract}
Feature matching is a crucial task in the field of computer vision, which involves finding correspondences between images. Previous studies achieve remarkable performance using learning-based feature comparison. However, the pervasive presence of matching redundancy between images gives rise to unnecessary and error-prone computations in these methods, imposing limitations on their accuracy. To address this issue, we propose MESA, a novel approach to establish precise \textbf{area} (or \textbf{region}) matches for efficient matching redundancy reduction. MESA first leverages the advanced image understanding capability of SAM, a state-of-the-art foundation model for image segmentation, to obtain image areas with implicit semantic. Then, a multi-relational graph is proposed to model the spatial structure of these areas and construct their scale hierarchy. Based on graphical models derived from the graph, the area matching is reformulated as an energy minimization task and effectively resolved. Extensive experiments demonstrate that MESA yields substantial precision improvement for multiple point matchers in indoor and outdoor downstream tasks, \eg~$+13.61\%$ for DKM in indoor pose estimation.
\end{abstract} 

\blfootnote{$^{*}~$Corresponding author: Xu Zhao.}

\section{Introduction}\label{sec:intro}

Feature matching aims at establishing correspondences between images, which is vital in a broad range of applications, such as SLAM~\cite{orbslam}, SfM~\cite{colmap} and visual localization~\cite{loc}.
However, achieving exact point matches is still a significant challenge due to the presence of {matching noises}~\cite{matchsurvey}, including scale variations, viewpoint and illumination changes, repetitive patterns, and poor texturing.

\begin{figure}[!t]
\centering
\includegraphics[width=\linewidth]{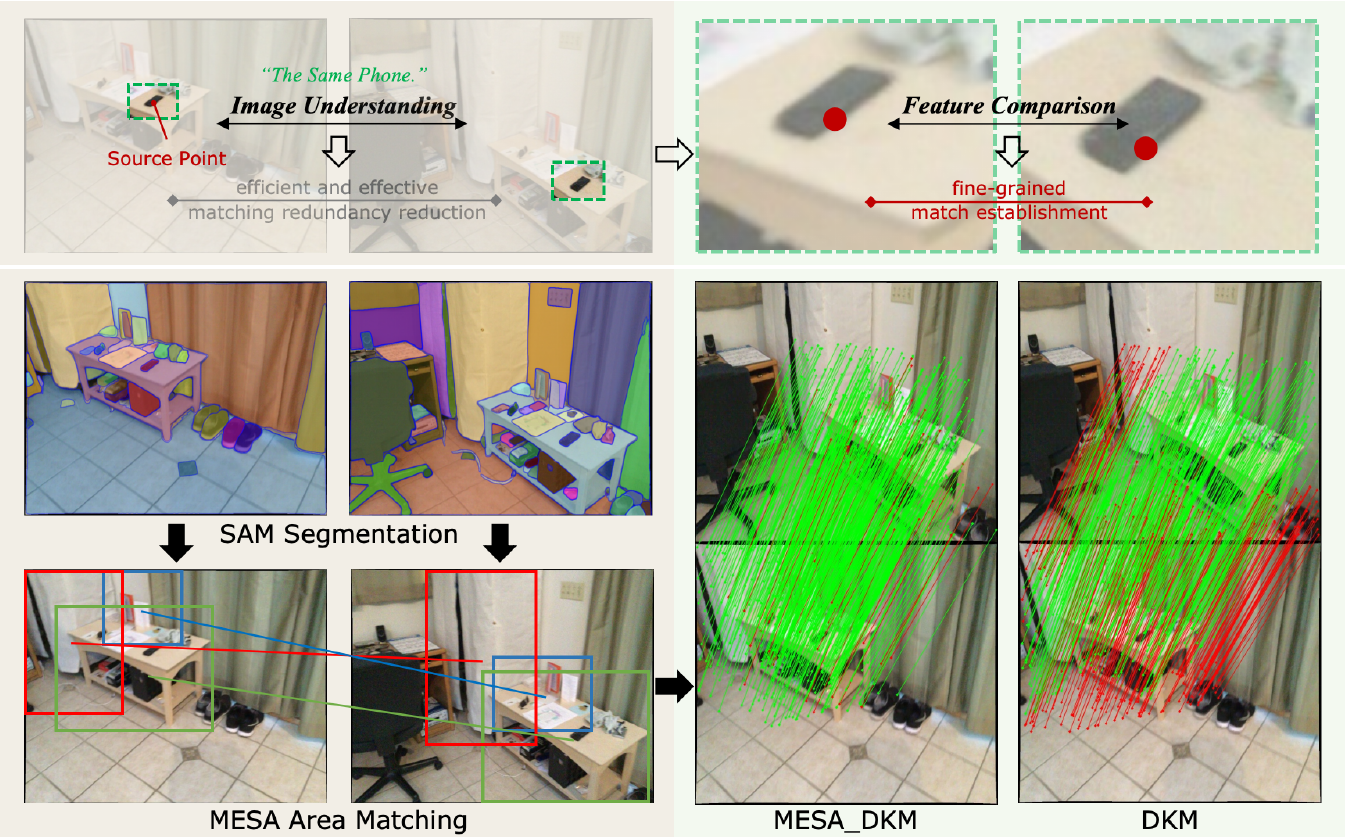}
\caption{\textbf{The matching redundancy reduction in MESA.} High-level 
\colorbox[rgb]{0.95,0.93,0.90}{image understanding} enables efficient matching redundancy reduction, allowing for precise point matching by dense \colorbox[rgb]{0.95,0.98,0.94}{feature comparison}. Therefore, MESA effectively reduces the matching redundancy by area matching based on SAM~\cite{sam} segmentation, significantly improving the accuracy of DKM~\cite{dkm}. }
\vspace{-1.5em}
\label{fig_first}
\end{figure}

The underlying goal of feature matching is to find correspondences for points in the source image from a highly redundant candidate set, \ie all the points of the target image. 
Therefore, feature matching essentially involves reducing the \textbf{matching redundancy}, which is caused by troublesome matching noises. 
Recent years have witnessed significant advancements in learning-based methods for feature matching. Especially the semi-dense~\cite{loftr,aspanformer} and dense methods~\cite{dkm}, which search for matches densely by deep feature comparison, obtain an impressive precision gap over sparse methods~\cite{superpoint}. 
However, as they heavily rely on dense feature comparison of the entire image, limited input resolution and error-prone computation in matching-redundant parts of images decrease their accuracy.
Thus, effective matching redundancy reduction is urgently required for them.

On the other hand, most matching redundancy can be effectively and efficiently identified through high-level image understanding and only strongly correlated local \textbf{areas} (also known as \textbf{regions}) need dense feature comparison to determine precise matches. (\cref{fig_first}).
Therefore, recent methods~\cite{adamatcher,OETR} perform learning-based overlap segmentation before matching. However, implicit learning needs non-reusable computational overhead and the redundancy still exists in the overlap. 
To address these issues, some work turn to explicit semantic prior~\cite{topicfm,sgam}.
Unlike manually specifying a topic number to group feature patches~\cite{topicfm}, \textit{Area to Point Matching} (A2PM) framework proposed by SGAM~\cite{sgam} provides a more intuitive way to reduce matching redundancy.
Specifically, this framework establishes semantic area matches first where matching redundancy is largely removed, and then obtains accurate correspondences within these areas utilizing an state-of-the-art point matcher.
Nevertheless, SGAM relies on semantic segmentation to determine area matches.
Its performance, thus, decreases when encountering inexact semantic labeling and semantic ambiguity~\cite{sgam}.
Furthermore, the benefits of A2PM cannot be extrapolated to more general scenes due to the close-set semantic labels.
Hence, reducing matching redundancy through semantic segmentation suffers from impracticality.

Recently, the Segment Anything Model (SAM)~\cite{sam} has gained notable attention from the research community due to its exceptional performance and versatility, which has the potential to be the basic front-end of many tasks~\cite{fasterSAM,midSAM,inpSAM}. This suggests that the foundation model can accurately comprehend image contents across various domains. Taking inspiration from this, we realize that the image understanding from the foundation model can be leveraged to reduce matching redundancy.
Thus, we propose to establish area matches based on SAM segmentation to overcome limitations of SGAM~\cite{sgam}.
Similar to the semantic segmentation, the image segmentation also provides multiple areas in images, but without semantic labels attached to these areas.
However, the general object perception of SAM ensures that its segmentation results inherently contain implicit semantic information. In other words, a complete semantic entity is always segmented as an independent area by SAM.
Hence, matching these areas effectively reduces matching redundancy and promotes accurate point matching within areas~\cite{sgam}.
Furthermore, the absence of explicit semantics alleviate issues of inaccurate area matching caused by erroneous labeling. The generic limitations due to semantic granularity are also overcome.
Nevertheless, area matching can not be simply achieved by semantic labels but requires other approaches under this situation.

\begin{figure*}[!t]
\centering
\includegraphics[width=\linewidth]{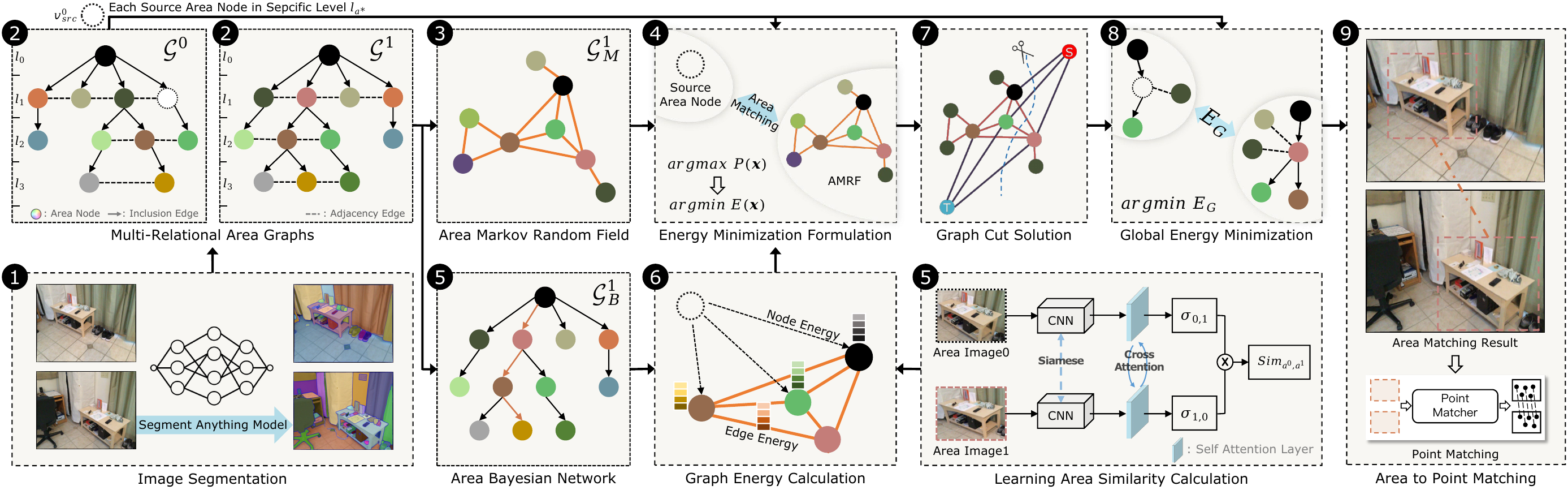}
\caption{\textbf{Overview of MESA.} Based on \ding{182}~\textit{SAM segmentation}, we first construct \ding{183}~\textit{Area Graph}s. Then the graph is turned to two graphical models based on its two different edges. Through \ding{184}~\textit{Area Markov Random Field}, area matching is formulated as an \ding{185}~\textit{Energy Minimization}. Then, leveraging \ding{186}~\textit{Area Bayesian Network} and our \ding{186}~\textit{Learning Area Similarity Calculation}, \ding{187}~\textit{Graph Energy} can be efficiently calculated. Therefore, \ding{188}~\textit{Graph Cut} is utilized to obtain putative area matches. Finally, \ding{189}~\textit{Global Energy Minimization} determines the best area match, which serves as the input of subsequent point matcher for precise feature matching, following the \ding{190}~\textit{Area to Point Matching} framework~\cite{sgam}.}
\vspace{-1.2em}
\label{fig_main}
\end{figure*}

In this work, we propose {Matching Everything by Segmenting Anything} (MESA, \cref{fig_main}), a method for precise area matching from SAM segmentation. 
MESA focuses on \textbf{two main aspects}: \textit{{constructing area relations}} and \textit{finding area matches based on these relations}.
Since SAM areas only provide local information, matching them independently can lead to inaccurate results, especially in scenes with scale variation and repetitiveness.
To address this, MESA first constructs a novel graph structure, named \textit{Area Graph} (AG), to model the global context of areas for subsequent precise matching.
The AG takes areas as nodes and connects them with two types of edges: undirected edges for adjacent areas and directed edges for areas that include one another.
Both edges capture global information, and the latter also enables the construction of hierarchy structures in AG similar to~\cite{quadtree} for efficient matching. 
Afterwards, MESA performs area matching by deriving two distinct graphical models from the AG: \textit{Area Markov Random Field} (AMRF) and \textit{Area Bayesian Network} (ABN).
The AMRF involves global-informative edges in AG, thus allowing global-consistent area matching through energy minimization on the graph.
The ABN, furthermore, is proposed to facilitate the graph energy calculation, through the hierarchy structure of AG.
Specifically, the graph energy is determined based on the similarity and spatial relation between areas, making this energy minimization effectively solvable through the \textit{Graph Cut}~\cite{GCE}.
The area similarity calculation is accomplished by a novel learning-based model and accelerated by the ABN.
Finally, we propose a global matching energy to refine accurate area matches, addressing the issue of multiple solutions in \textit{Graph Cut}, leading to effective matching redundancy reduction.

Our method makes several contributions. \textbf{1)} In pursuit of effective matching redundancy reduction, we propose MESA to achieve accurate area matching from results of SAM, an advanced image segmentation foundation model. MESA realizes the benefit of A2PM framework in a more practical manner, leading to precise feature matching. \textbf{2)} We introduce a multi-relational graph, AG, to model the spatial structure and scale hierarchy of image areas, contributing to outstanding area matching. \textbf{3)} We propose graphical area matching by converting AG into AMRF and ABN. The AMRF is employed to formulate the area matching as an energy minimization, which is effectively solvable through the \textit{Graph Cut}. The energy calculation is efficiently performed by the property of ABN and the proposed learning area similarity model. \textbf{4)} MESA is evaluated in indoor and outdoor downstream tasks of feature matching, obtaining remarkable improvements for multiple point matchers.

%
\section{Related Work}
\boldparagraph{Sparse, Semi-Dense and Dense Matching.}
There are three types of feature matching methods: sparse, semi-dense and dense.
Classical sparse feature matching~\cite{sift} involves detecting and describing keypoints in images, followed by keypoint matching.
The learning counterpart of this framework utilizes neural networks to perform feature detection~\cite{keyNet,superpoint}, description~\cite{d2net,r2d2,alike} or matching~\cite{superglue,OANet}.
To avoid the detection failure in sparse methods, semi-dense methods~\cite{loftr,cotr,DRC-Net} are proposed, also known as the detector-free methods.
These methods~\cite{aspanformer,quadtree,ecotr} perform dense feature matching over the entire image and then select confident matches, which achieve impressive matching precision.
Dense matching methods~\cite{dkm,dgc,glu} output a dense warp with confidence map for the image pair. Recent DKM~\cite{dkm} reformulates this framework as a Gaussian Process and achieves \textit{state-of-the-art} performance.
Our method, however, focus on area matching between images, which can be combined with both semi-dense and dense point matching methods to increase their precision.


\boldparagraph{Matching Redundancy Reduction.}
Matching redundancy is evident in non-overlapping areas between images, motivating several works~\cite{cotr,OETR,oamatcher} to focus on covisible areas extraction. They predict overlaps between images by iterative matching~\cite{cotr} or overlap segmentation~\cite{OETR,oamatcher}. Nevertheless, matching redundancy still exists in the overlapping area, when it comes to detailed point matching. TopicFM~\cite{topicfm} proposes to divide image contents into topics and then restrict matching to the same topic to avoid redundant computation. The learning-based topic inference, however, is implicit, suffering from lack of versatility and generalize issues. SGAM~\cite{sgam} explicitly finds area matches between images, which are fed into point matchers for redundancy reduction. The Area to Point Matching framework is simple yet effective. Our method further builds on its advantages.

\boldparagraph{Area to Point Matching.}
Establishing area matches is an effective way to reduce the matching redundancy.
Thus, SGAM~\cite{sgam} proposes the Area to Point Matching (A2PM) framework.
This framework obtains area matches as the input of point matchers, which possess higher resolution and less matching noise than original images, contributing to accurate point matching.
However, SGAM heavily relies on explicit semantic prior, leading to performance sensitivity of area matching to semantic labeling precision and scene semantic granularity.
In contrast, MESA utilizes pure image segmentation for area matching, which is more practical and remedies drawbacks associated with explicit semantic.

\section{Area Graph}\label{sec:ag}
In this section, we introduce the concept of an Area Graph (AG) and explain how to construct the AG from the SAM~\cite{sam} results of an image.
The main reason to adopt the AG is that direct area matching on SAM results is unacceptable, as global information is ignored in independent areas.
Also, fixed area sizes hinder robust point matching under scale changes.
Therefore, AG is proposed to capture the global structural information of these areas and construct scale hierarchy for them, contributing to accurate and robust area matching.

\subsection{Area Graph Definition}
\begin{figure}[!t]
\centering
\includegraphics[width=\linewidth]{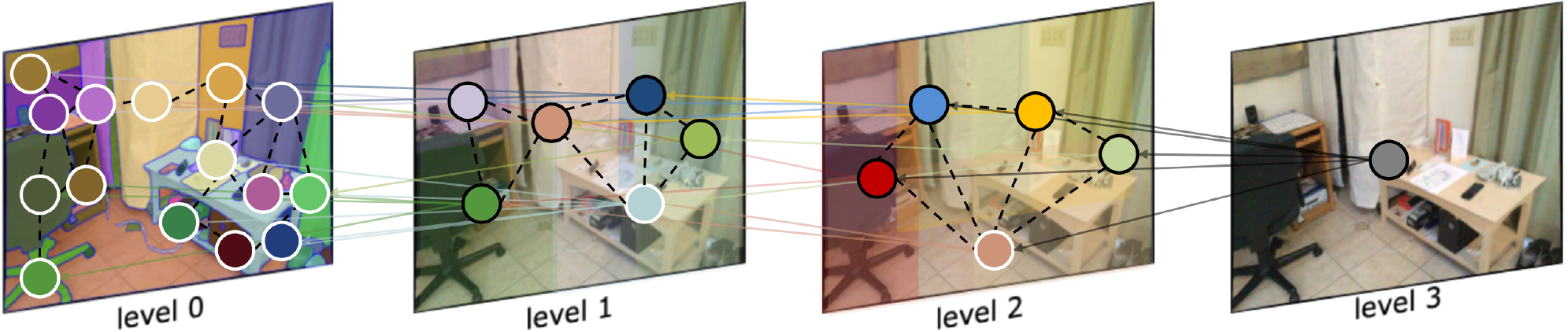}
\caption{\textbf{The proposed Area Graph.} The graph nodes (circles with masks representing rectangle areas) includes both areas from SAM results (white boundaries) and our graph completion algorithm (black boundaries). They are divided into four levels according to their sizes. The adjacency edges (dashed lines) and inclusion edges (arrows) connect all nodes. Only adjacency edges within the same level are shown for better view.}
\vspace{-1.5em}
\label{fig:ag}
\end{figure}

The AG ($\mathcal{G}\!=\!\langle \mathcal{V}, \mathcal{E}\rangle$) takes image areas as nodes and contains two kinds of edges to model inter-area relations (\cref{fig:ag}), thus making it a \textit{multi-relational graph}~\cite{mrg}.
The graph nodes include both areas provided by SAM and areas generated by graph completion (cf.~\cref{set:gc}).
We divide these areas into $L$ levels according to their sizes, corresponding to different image scales, to serve as the foundation of area scale hierarchy.
On the other hand, the graph edges ($\mathcal{E}\! = \!\mathcal{E}_{in} \bigcup \mathcal{E}_{adj}$) represent two relations between areas, \ie inclusion ($\mathcal{E}_{in}$) and adjacency ($\mathcal{E}_{adj}$).
The inclusion edge $e_{in}\!\in\!\mathcal{E}_{in}$ is directed, pointing from an area to one of its containing areas.
It forms a hierarchical connection between graph nodes, enabling robust and efficient area matching especially under scale changes.
The adjacency edge $e_{adj}\!\in\!\mathcal{E}_{adj}$ is undirected, indicating the areas it connects share common parts but without the larger one including the smaller one.
This edge captures the spatial relations between areas.
By the above two edges, AG models both the spatial and scale structure of image areas, thus contributing to exact area matching.

\subsection{Area Graph Construction}\label{set:gc}
The construction of AG includes collecting areas as nodes and connecting them by proper edges.
Firstly, not all SAM areas can function as nodes, since some are too small or have extreme aspect ratios, rendering them unsuitable for point matching. 
Thus, \textit{Area Pre-processing} is performed first to obtain initial graph nodes.
We then approach the edge construction as a \textit{Graph Link Prediction} problem~\cite{lpSurvey}.
Afterwards, the preliminary AG is formed, but it still lacks matching efficiency and scale robustness.
Thus, we propose the \textit{Graph Completion} algorithm, which generates additional nodes and edges to construct the scale hierarchy. 

\boldparagraph{Area Pre-processing.}
To filter unsuitable areas, we set two criteria: the acceptable minimal area size ($T_s$) and maximum area aspect ratio ($T_r$).
Any area that has smaller size than $T_s$ or larger aspect ratio than $T_r$, gets screened out.
The remaining areas are added into the candidate areas set.
For each filtered area, we fuse it with its nearest neighbor area in the candidate set.
We repeat the filtering and fusion on the candidate set until no areas get screened out.
Then, we assign a level $l_a$ to each candidate area $a$ based on its size, by setting $L$ size thresholds ($\{TL_i~\big|~i\in[0,L-1]\}$):
\begin{equation}
    l_a = i \; \big| \; TL_{i} \leq W_a \times H_a < TL_{i+1}.
\end{equation}
The size level is the basis of scale hierarchy in AG.

\boldparagraph{Graph Link Prediction.}
The edge construction is treated as a link prediction problem~\cite{lpSurvey}.
Given two area nodes ($v_i,v_j$), the edge between them ($e_{ij}$) can be predicted according to the spatial relation of their corresponding areas ($a_i, a_j$).
This approach adopts the ratio of \textit{the overlap size} ($O_{ij}$) to \textit{the minimum area size} between two areas ($\delta = {O_{ij}}/{\min (W_i \times H_i,W_j \times H_j )}$) as the score function:
\begin{equation}\label{eq:lp}
e_{ij} \in \left\{
	\begin{aligned}
	&\mathcal{E}_{in} &,\; &\delta >= \delta_h \\
	&\mathcal{E}_{adj} &,\;&\delta_l < \delta < \delta_h \\
	 &\varnothing  &,\; &\delta \leq \delta_l\\
	\end{aligned}
	\right
	.
    ,
\end{equation}
where $\delta_l, \delta_h$ are predefined thresholds.

\boldparagraph{Graph Completion.}
Initial AG is achieved by connecting all processed nodes with different edges.
However, since SAM inherently produces areas containing complete entity, there are few inclusion relations among areas.
Consequently, initial AG lacks the scale hierarchy, which reduces its robustness at scale variations and makes accessing nodes inefficient.
To address the issue, we propose the \textit{Graph Completion} algorithm, which generates additional nodes and edges to ultimately construct a tree structure in the original graph.
The core of this algorithm is to generate parent nodes for each {orphan node}, \ie~the node has no parent node in the next higher level, from small to large size levels.
The algorithm begins at the smallest level, collects all orphan nodes, and clusters them based on their center coordinates.
Nodes in the same cluster have their corresponding areas fused with their nearest neighbors. It is noteworthy that these generated areas containing multiple objects preserve the internal implicit semantic. 
Based on our level thresholds, the resulting areas correspond to new higher level nodes.
If a node remains single after clustering, we increase its area size to the next level to allow for potential parent nodes.
We repeat the above operations on each level and connect generated nodes to others by suitable edges.
More details can be found in the Supp. Mat.

\section{Graphical Area Matching}
In this section, we describe our area matching method, which formulates area matching on the graph, utilizing two graphical models derived from the AG.
Given two AGs ($\mathcal{G}^0, \mathcal{G}^1$) of the input image pair ($I_0, I_1$) and one area ($a^0_{src}\!\in\!I_0$) corresponding to node $v^0_{src}\!\in\!\mathcal{G}^0 $ (termed as the \textit{source node}), area matching involves finding the node $v^1_j\!\in\!\mathcal{G}^1 $ with the highest probability of matching its corresponding area $a^1_j$ to the source node area $a^0_{src}$.
However, treating this problem as independent node matching is inadequate, as which disregards global information from adjacency and inclusion relations between areas modeled in AG by its two edges.
Meanwhile, these two edges respectively convert AG into two graphical models, \ie Markov Random Fields (undirected edges) and Bayesian Network (directed edges).
Thus, the area matching can be naturally formulated inside a framework of graphical model.

\subsection{Area Markov Random Field}
The adjacency edge in AG implies the correlation between probabilities of two nodes matching the source node.
By considering the general adjacency relation, which includes the inclusion relations as adjacency too, the $\mathcal{G}^1$ is transformed into an undirected graph.
Next, random variables ($\bm{x}$) are introduced for all nodes to indicate their matching status with the source node.
The binary variable $x_i\in \bm{x}$ is equal to $1$ when $v^1_i$ matches $v^0_{src}$ and $0$ otherwise.
Therefore, Area Markov Random Field (AMRF, $\mathcal{G}^1_M = \langle \mathcal{V}, \mathcal{E}_{adj} \rangle$) is obtained and area matching can be performed by maximizing the joint probability distribution over the AMRF:
\begin{equation}
    \arg\max_{\bm{x}} P(\bm{x}).
\end{equation}
Based on the \textit{Hammersley-Clifford} theorem~\cite{markov}, the probability distribution defined by AMRF belongs to the \textit{Boltzmann distribution}, which is an exponential of negative energy function ($P(\bm{x})\!=\!\exp(-E(\bm{x}))$). 
Therefore, the area matching can be formulated as an energy minimization.
\begin{equation}\label{eq:em}
    \arg\min_{\bm{x}} E(\bm{x}). 
\end{equation}
The energy can be divided into two parts, \ie the energy of nodes ($E_{\mathcal{V}}$) and edges ($E_{\mathcal{E}}$), based on the graph structure.
\begin{equation}\label{eq:e}
    E(\bm{x})=\sum_iE_{\mathcal{V}}(x_i)+\lambda\sum_{(i,j)\in\mathcal{N}}E_{\mathcal{E}}(x_i,x_j),
\end{equation}
where $\lambda$ is a parameter balancing the terms and $\mathcal{N}$ is the set of all pairs of neighboring nodes.
For each graph node $v^1_i$, its energy is expected to be low when its matching probability is high, which can be reflected by the apparent similarity ($S_{a^0_{src}a^1_i}$) between $a^0_{src}$ and $a^1_i$.
\begin{equation}
    E_{\mathcal{V}}(x_i)=|x_i-S_{a^0_{src}a^1_i}|.
\end{equation}
The edge energy aims to penalize all neighbors with different labels, and the Potts model~\cite{potts} ($T$) would be a justifiable choice.
To better reflect the spatial relation, the Potts interactions are specified by $IoU$~\cite{IoU} of neighboring areas.
\begin{equation}
    E_{\mathcal{E}}(x_i,x_j)=IoU(a^1_i,a^1_j) \cdot T(x_i\not=x_j).
\end{equation}
Function $T(\cdot)$ is $1$ if the argument is true and $0$ otherwise.
Finally, the area matching is formulated as an binary labeling energy minimization.
By carefully defining the energy function, the energy minimization problem in \cref{eq:em} is efficiently solvable via the \textit{Graph Cut} algorithm~\cite{GC,GCE}.
The obtained minimum cut of the graph $\mathcal{G}^1_M$ is the matched node set ($\{v^1_{h}\big | h \in {\mathcal{H}}\}$).
Although the set may contain more than one area node, the best matching result can be achieved from this set by our refinement algorithm (cf. \cref{Sec:refine}).

\begin{figure}[!t]
\centering
\includegraphics[width=\linewidth]{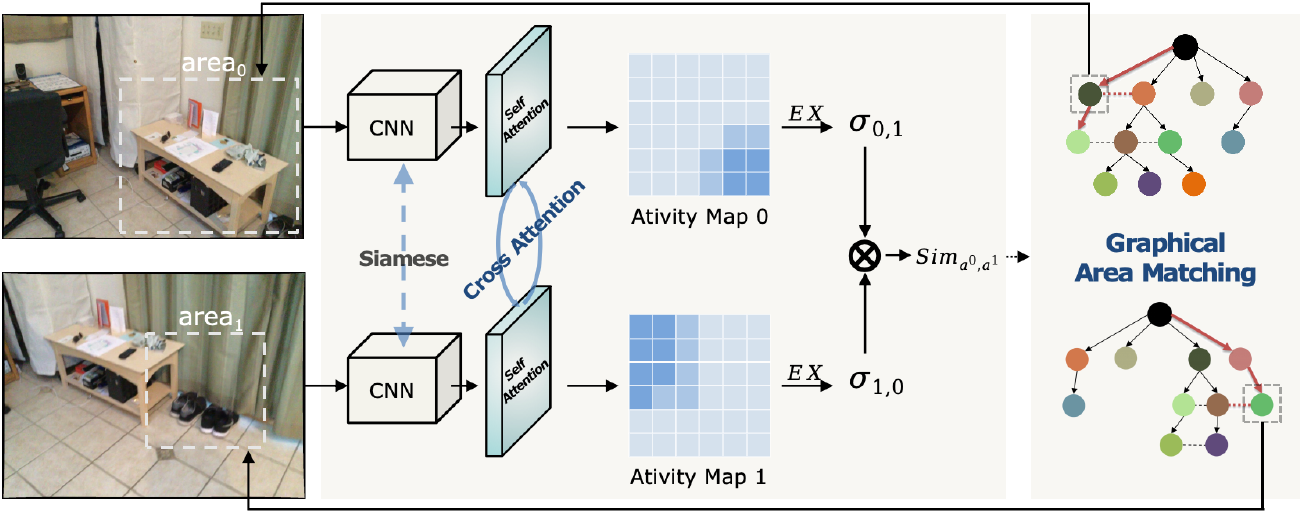}
\caption{\textbf{Learning area similarity.} The area similarity calculation is formed as the patch-level classification. We predict the probability of each patch in one area appearing on the other to construct activity maps. The similarity is obtained by the product of activity expectations, contributing to our exact area matching.}
\vspace{-1.2em}
\label{fig:las}
\end{figure}

\subsection{Learning Area Similarity}\label{sec:las}
The proposed \textit{Graph Cut} solution relies on graph energy calculations for both nodes and edges. 
Unlike easily available area pair \textit{IoU}s for $E_{\mathcal{E}}$, determining the area apparent similarity for $E_{\mathcal{V}}$ is not straightforward.
Thus, we turn to the learning-based framework, inspired by recent successes of learning models in point matching~\cite{aspanformer,loftr,pats}.
One simple idea is to calculate the correlation of learning descriptors of two areas~\cite{pats} as the area similarity.
However, the descriptor correlation is too rough for accurate area matching and lacks fine-grained interpretability.
To overcome these issues, we decompose the area similarity calculation into two patch-level classification problems as shown in \cref{fig:las}.

Specifically, for each image in the area image pair $\{a_j\big|j\in\{0,1\}\}$ reshaped to the same size, we perform binary classification for each $1/8\cdot  W_{a_j}\times1/8\cdot H_{a_j}$ image patch $p^j_i$ (\textit{where $i$ is the index of patch and $j$ is the index of area image}) in it, computing the probability of $p^j_i$ appearing on the other area image, termed as the patch activity $\sigma^j_i$.
To accomplish the classification, we first extract patch-wise features from each area image using a Siamese CNN~\cite{siamese}.
Then we update these patch features via self and cross-attention with normalization~\cite{attention}, resulting in patch activities.
Utilizing these patch activities, we construct an activity map ($\sigma^j_m = \{\sigma^j_i\big | j \in \{0,1\}\}_i$, \textit{where $i$ is the index of element in this set}) for each area image.
When two areas are ideally matched, the corresponding 3D point of every pixel in one area is projected onto the other area.
Hence, all the patch activities of both areas should be closed to $1$, revealing the area similarity can be represented by the product of expectations of two activity maps, $EX(\sigma_m^i)=\sigma_{i,j}$, where $ i\not=j$, $i,j \in \{0,1\}$.
\begin{equation}
    Sim_{a^0,a^1} = EX(\sigma_m^0) \times EX(\sigma_m^1) = \sigma_{0,1} \times \sigma_{1,0}.
\end{equation}
The training details of this learning model are described in the Supp. Mat.

\subsection{Area Bayesian Network}\label{sec:abn}
Although the \textit{Graph Cut} can be performed in polynomial time~\cite{GC}, the dense energy calculation over $\mathcal{G}_M^1$ is time consuming.
Furthermore, due to the scale hierarchy in AG, this dense calculation is highly redundant.
In particular, if the area $a^0_{src}$ is not matched to $a^1_j$, it won't be matched to any children area of $a^1_j$.
This observation reveals the conditional independence in the similarity calculation, which involves inclusion edges in $\mathcal{G}^1$, thus turning $\mathcal{G}^1$ to a Bayesian Network ($\mathcal{G}^1_B$)~\cite{prml}.
Therefore, the redundancy in the similarity calculation can be reduced.
In practice, we calculate the dense similarities by constructing a similarity matrix $M_S\!\in\!\mathbb{R}^{|\mathcal{V}^0| \times |\mathcal{V}^1|}$. Note \textit{not} all similarities in $M_S$ need calculation, but any similarity can be accessed in $M_S$. We first calculate similarities related to all source nodes. Subsequent calculations are saved in $M_S$ as well.
For $M_S[i,j]$ that has not been calculated, we achieve it by our learning model (cf. \cref{sec:las}):
\begin{equation}
    M_S[i,j] = Sim_{a^0_{i}, a^1_j}.
\end{equation}
If $M_S[i,j] < T_{as}$, all children nodes $\{v^0_{h}\big | h\in ch^0(i)\}$ and $\{v^1_{c}\big | c\in ch^1(j)\}$ of $v^0_{i}$ and $v^1_j$ are found from $\mathcal{G}_B^0$ and $\mathcal{G}_B^1$, where $ch^j(i)$ is the index set of children indices of node $v_i^j$ from $\mathcal{G}^j_B$.
Based on the conditional independence, we have:
\begin{equation}\label{eq:abn}
    M_S[h,k] = 0,~\forall (h,k) \in ch^0(i) \times ch^1(j).
\end{equation}
This operation effectively reduce the times of similarity calculation, leading to more efficient area matching.

\subsection{Global Matching Energy Minimization}\label{Sec:refine}
The minimum cut $\{v^1_{h} \big | h \in {\mathcal{H}}\}$ achieved through the \textit{Graph Cut} may contain more than one area node, which means further refinement is necessary to obtain the best area match.
Moreover, the aforementioned graphical area matching, \ie finding the corresponding area node in $\mathcal{G}^1$ for $v^0_{src} \in \mathcal{G}^0 $, only considers the structure information in $\mathcal{G}^1$ and ignores the structure of $\mathcal{G}^0$.
To overcome this limitation, we propose a global matching energy $E_G$ for each candidate node $v^1_{h}$, which consists of four parts.
\begin{equation}\label{eq:eg}
    \begin{aligned}
       E_G(v^1_{{h}}) & = \frac{1}{Z}( \mu \cdot E_{self}(v^1_{{h}})+\alpha  \cdot  E_{parent}(v^1_{{h}}) \\ 
         & +\beta \cdot E_{children}(v^1_{{h}})+\gamma \cdot E_{neighbour}(v^1_{{h}}))
    \end{aligned},
\end{equation}
where $\mu, \alpha, \beta$ and $\gamma$ are weights to balance the terms; $Z$ is the partition function.
The $E_{self}(v^1_{{h}})$ is the energy related to matching probability between $v^0_{src}$ and $v^1_{{h}}$:
\begin{equation}\label{eq:e_self}
    E_{self}(v^1_{{h}}) = |1 - Sim_{a^0_{src},a^1_h}|.
\end{equation}
The $E_{parent}(v^1_{{h}})$ is the energy related to matching probability between the parent node pairs of $v^0_{src}$ and $v^1_{{h}}$:
\begin{equation}
\centering
\begin{gathered}
         E_{parent}(v^1_{{h}}) = \\ 
         \min \{|1 - Sim_{a^0_{u},a^1_{r}}| \big| \small u \in p^0(src), r \in p^1(h)\},
\end{gathered}
\end{equation}
where $p^i(j)$ is the index set of parent nodes of $v^i_j$ in $\mathcal{G}_i$.
This energy is the minimum matching energy among all parent node pairs of $v^1_{{h}}$ and $v^0_{src}$.
Same as the $E_{parent}$, the $E_{children}$ and $E_{neighbour}$ are the energy parts of children and neighbour node matching pairs.
Afterwards, the best area match $v^1_{h^*}$ in the set can be found by minimize $E_G$:
\begin{equation}
     h^* = \arg\min_{h \in {\mathcal{H}}} E_G(v^1_{h}).
\end{equation}
If the $ E_G(v^1_{h^*}) > T_{E_{max}}$ (a threshold parameter), the source area node $v^0_{src}$ is considered to have no matches.
To further improve the accuracy of final match, we set an energy range threshold $T_{Er}$ to collect all the candidates within a certain energy range.
\begin{equation}\label{eq:collectarea}
    \{v^1_{\bar{h}}\big||E_G(v^1_{\bar{h}}) -  E_G(v^1_{h^*})| \leq T_{Er}, \bar{h} \in {\mathcal{H}} \}.
\end{equation}
Then the final area match is achieved by fusing $v^1_{h^*}$ and all candidates $\{v^1_{\bar{h}}\}_{\bar{h}}$ with $E_G$ as weights, similar to~\cite{pats}.
This refinement completely considers the structure information of both $\mathcal{G}^0$ and $\mathcal{G}^1$ and achieves exact area matches.
\section{Experiments}

\begin{table}[t]
\resizebox{\linewidth}{13mm}{
\begin{threeparttable}
\begin{tabular}{rcccccc}
\toprule
\multicolumn{2}{c}{Method} & AOR $\uparrow$    & AMP@$0.6$ $\uparrow$ & AMP@$0.7$ $\uparrow$ & AMP@$0.8$ $\uparrow$ & AreaNum 
 $\uparrow$ \\ \cmidrule(r){1-2} \cmidrule(r){3-3} \cmidrule(r){4-6} \cmidrule(r){7-7}
\multicolumn{2}{c}{SEEM~\cite{SEEM}+SGAM~\cite{sgam}\tnote{$\dagger$}}             &  52.96 &  49.39 &  29.88 & \rd 12.32 & \fs 2.66      \\
\multicolumn{2}{c}{\small w/ \textit{GAM} ($\phi=0.3$)}  & \rd 57.74 & \rd 53.86 & \rd 33.61 & \nd 13.13 &  1.21      \\
\multicolumn{2}{c}{\small w/ \textit{GAM} ($\phi=1.0$)}  & \fs 60.59 & \fs 59.43 &  \nd 36.29 & \fs 15.27 &  \rd 1.55      \\
\multicolumn{2}{c}{\small w/ \textit{GAM} ($\phi=5.0$)}  & \nd 60.32 & \nd 59.34 & \fs 36.55 & \fs 15.27 & \nd 2.23      \\ \hdashline \noalign{\vskip 1pt}
\multicolumn{2}{c}{\textbf{MESA}}              &  67.98 &  80.09 &  57.74 &  22.73 & \fs{3.47}      \\
\multicolumn{2}{c}{\small w/ \textit{GAM} ($\phi=0.3$)}  & \fs{68.99} & \fs{81.11} & \fs{60.75} & \fs{25.77} &  1.97        \\
\multicolumn{2}{c}{\small w/ \textit{GAM} ($\phi=1.0$)}  & \nd 68.78 & \nd 81.07 & \nd 60.06 & \nd 25.21 & \rd 3.25      \\
\multicolumn{2}{c}{\small w/ \textit{GAM} ($\phi=5.0$)}  & \rd 68.51 & \rd 80.84 & \rd 59.22 & \rd 23.87 & \nd 3.35      \\ 
 \bottomrule
\end{tabular}
\begin{tablenotes}
     \item[$\dagger$] {\small SGAM with only semantic area matching activated.}
\end{tablenotes}
\end{threeparttable}
}
\vspace{-0.8em}
\caption{\textbf{Area matching results on ScanNet1500.} We compare the area matching performance between SGAM and MESA, combined with GAM~\cite{sgam} under various $\phi$ settings. Results are highlighted as \colorbox{colorFst}{\bf first}, \colorbox{colorSnd}{second} and \colorbox{colorTrd}{third} for both series respectively.}\label{tab:AM}
\vspace{-1.2em}
\end{table}

\subsection{Area Matching} \label{exp:am}
Since accurate area matching is the prerequisite for the precise point matching, we evaluate our method for this task on ScanNet1500~\cite{scannet} benchmark.

\boldparagraph{Experimental setup.} 
The $T_{E_{max}}$ is $0.35$. 
The input size is $320 \times 320$ for our learning model. Other scene-independent parameter settings and implementation details can be found in our Supp.~Mat. The previous area matching approach, SGAM~\cite{sgam} with SEEM~\cite{SEEM} providing the semantic segmentation input, is compared with our method MESA. The impacts of \textit{Geometric Area Matching} (GAM) in~\cite{sgam} to both methods are also investigated, whose parameter $\phi$ reflects the strictness of outlier rejection. We employ the \textit{Area Overlap Ratio} (AOR, \%) and \textit{Area Matching Precision} under three thresholds (AMP@$0.6/0.7/0.8$, \%)~\cite{sgam} as metrics. The average area number per image (AreaNum) is reported.

\boldparagraph{Results.}
In \cref{tab:AM}, our MESA outperforms SGAM by a large margin ($67.98~vs.~60.59$ in AOR). 
The poor area matching performance of SGAM is mainly caused by the inaccurate semantic labeling of SEEM, which is avoided in MESA.
MESA can also achieve more areas than SGAM, favouring downstream tasks by more point matches. Although GAM brings significant improvement for SGAM, it achieves less enhancement on MESA, because more precise area matching is obtained by MESA. We empirically adopt GAM w/ $\phi=1.0$ for MESA in the following experiments.

\begin{table}[t]
\centering
\resizebox{\linewidth}{30mm}{
\begin{tabular}{lllllll}
\toprule
\multicolumn{3}{l}{\multirow{3}{*}{Pose estimation AUC}} & \multicolumn{3}{c}{ScanNet1500 benchmark} \\ \cmidrule(l){4-6} 
\multicolumn{3}{c}{}                        & AUC@5$^\circ\uparrow$       & AUC@10$^\circ\uparrow$        & AUC@20$^\circ\uparrow$        \\ \midrule
\multicolumn{3}{l}{SP~\cite{superpoint}+SGMNet~\cite{SGMNet}~\tiny{ICCV'21}}               & 15.4      & 32.1       & 48.3       \\
\multicolumn{3}{l}
{SiLK~\cite{silk}~\tiny{ICCV'23}}               & {18.0}      & 34.4      & 50.4    \\
 \midrule
\multicolumn{3}{l}{PATS~\cite{pats}~\tiny{CVPR'23}}                    & {26.0}       & {46.9}       & {64.3}       \\ 
\multicolumn{3}{l}
{CasMTR~\cite{casmtr}~\tiny{ICCV'23}}               & 27.1      & 47.0      & {64.4}     \\
\midrule
\multicolumn{3}{l}{ASpan~\cite{aspanformer}~\tiny{ECCV'22}}                   & 25.6       & 46.0       & 63.3       \\
\multicolumn{3}{l}{ SEEM~\cite{SEEM}+SGAM\_ASpan~\cite{sgam}~\tiny{arxiv'23}}        & {27.5}$_{+7.42\%}$      & {{48.0}}$_{+4.35\%}$      & {65.3}$_{+3.16\%}$       \\ 
\rowcolor[rgb]{.9,.95,.98}\multicolumn{3}{l}{MESA\_ASpan}        & {27.5}$_{+7.42\%}$      & {48.2}$_{+4.35\%}$      & {65.0}$_{+2.69\%}$       \\\midrule
\multicolumn{3}{l}{QuadT~\cite{quadtree}~\tiny{ICLR'22}}                   & 24.9       & 44.7       & 61.8       \\
\multicolumn{3}{l}{SEEM~\cite{SEEM}+SGAM\_QuadT~\cite{sgam}~\tiny{arxiv'23}}        & 25.5$_{+2.53\%}$       & 46.0$_{+2.95\%}$       & 63.4$_{+2.59\%}$       \\ 
\rowcolor[rgb]{.9,.95,.98}\multicolumn{3}{l}{MESA\_QuadT}        & 28.7$_{+15.26\%}$       & 46.5$_{+4.03\%}$       & 61.9$_{+0.16\%}$       \\ \midrule
\multicolumn{3}{l}{LoFTR~\cite{loftr}~\tiny{CVPR'21}}                   & 22.1       & 40.8       & 57.6       \\
\multicolumn{3}{l}{SEEM~\cite{SEEM}+SGAM\_LoFTR~\cite{sgam}~\tiny{arxiv'23}}        & 23.4$_{+5.88\%}$       & 41.8$_{+2.45\%}$       & 58.7$_{+1.91\%}$       \\ 
\rowcolor[rgb]{.9,.95,.98}\multicolumn{3}{l}{MESA\_LoFTR}        & 22.9$_{+3.62\%}$       & 41.8$_{+2.45\%}$       & 58.4$_{+1.39\%}$       \\ \midrule
\multicolumn{3}{l}{DKM~\cite{dkm}~\tiny{CVPR'23}}                   & 29.4       & {50.7}       & {68.3}       \\
\multicolumn{3}{l}{SEEM~\cite{SEEM}+SGAM\_DKM~\cite{sgam}~\tiny{arxiv'23}}        & \uline{30.6}$_{+4.12\%}$       & \uline{52.3}$_{+3.10\%}$       & \uline{69.3}$_{+1.48\%}$       \\ 
\rowcolor[rgb]{.9,.95,.98}\multicolumn{3}{l}{MESA\_DKM}        & \textbf{33.4}$_{+13.61\%}$       & \textbf{55.0}$_{+8.48\%}$       & \textbf{72.0}$_{+5.42\%}$       \\ 
\bottomrule
\end{tabular}
}
\vspace{-0.5em}
\caption{\textbf{Relative pose estimation results (\%) on ScanNet1500 benchmark.}  The \textbf{best} and \uline{second} results are highlighted.}\label{tab:PAUC_SN}
\vspace{-1.2em}
\end{table}

\subsection{Indoor Pose Estimation}
MESA is also evaluated using the ScanNet1500 benchmark for indoor pose estimation. The benchmark is challenging due to the poor texture and large view variances in scenes.

\boldparagraph{Experimental setup.}
The parameters of MESA are the same as our area matching experiment in \cref{exp:am}. We adopt the A2PM framework~\cite{sgam} and combine four fine-tuned point matchers with our method, \ie LoFTR~\cite{loftr}, ASpan~\cite{aspanformer}, QuadT~\cite{quadtree} and DKM~\cite{dkm}, to demonstrate our performance and compare with recent methods~\cite{silk,casmtr,pats}. The cropped area images are resized to be square with the same side length as the long side of the default size of point matchers. Following~\cite{loftr}, the pose estimation AUC at various thresholds are used as metrics. More implementation details can be found in our Supp. Mat.

\boldparagraph{Results.}
As we can see in \cref{tab:PAUC_SN}, MESA significantly increases the pose accuracy for all baselines.
Due to the presence of more severe matching redundancy in indoor scenes, our approach gains more noticeable enhancement, \eg~${+15.26\%}$ in AUC@$5^\circ$ for QuadT, compared to outdoor scenes in \cref{tab:PAUC_MD}.
The improvement obtained by MESA for semi-dense matchers are comparable to those of SGAM~\cite{sgam}, because the coarse-to-fine pipeline of these methods makes them insensitive to AOR.
However, MESA brings a larger improvement for DKM~\cite{dkm} than SGAM ($13.61\%~vs.~4.12\%$) and boosts the precision to a new \textit{state-of-the-art}, proving the precise area matching by MESA benefits the performance of dense matching.

\subsection{Outdoor Pose Estimation}
We use the Megadepth1500~\cite{megadepth} which consists of 1500 outdoor image pairs to evaluate the effectiveness of our method for relative pose estimation in outdoor scenes.

\boldparagraph{Experimental setup.} The parameter $T_{E_{max}}$ is set as $0.3$. MESA is combined with four baselines to compare with recent methods~\cite{astr,lightglue,topicfm}. We also compare with another covisible-area matching method, OETR~\cite{OETR}, combined with our post-processing. The area images are cropped and resized as input of point matcher, same as the indoor experiment. The pose estimation AUC is adopted as the metric.

\begin{table}[t]
\centering
\resizebox{\linewidth}{!}{
\begin{tabular}{ccclll}
\toprule
\multicolumn{3}{l}{\multirow{3}{*}{Pose estimation AUC}} & \multicolumn{3}{c}{MegaDepth1500 benchmark} \\ \cmidrule(l){4-6} 
\multicolumn{3}{c}{}                        & AUC@5$^\circ\uparrow$       & AUC@10$^\circ\uparrow$        & AUC@20$^\circ\uparrow$        \\ \midrule
\multicolumn{3}{l}
{LightGlue~\cite{lightglue}~\tiny{ICCV'23}}               & {49.9}      & 67.0      & 80.1    \\
\multicolumn{3}{l}
{SiLK~\cite{silk}~\tiny{ICCV'23}}               & {43.8}      & 57.7      & 68.6    \\
\midrule
\multicolumn{3}{l}{ASTR~\cite{astr}~\tiny{CVPR'23}}                    & {58.4}       & {73.1}       & {83.8}       \\
\multicolumn{3}{l}{TopicFM~\cite{topicfm}~\tiny{AAAI'23}}                    & {54.1}       & {70.1}       & {81.6}       \\ 
\multicolumn{3}{l}
{CasMTR~\cite{casmtr}~\tiny{ICCV'23}}               & 59.1      & 74.3      & {84.8}     \\
\midrule
\multicolumn{3}{l}{ASpan~\cite{aspanformer}~\tiny{ECCV'22}}                   & 55.4       & 71.6      & 83.1       \\
\multicolumn{3}{l}{OETR~\cite{OETR}+ASpan~\tiny{AAAI'22}}        & {56.2}$_{+1.44\%}$      & {72.3}$_{+0.98\%}$      & {82.8}$_{-0.36\%}$       \\
\rowcolor[rgb]{.9,.95,.98}\multicolumn{3}{l}{MESA\_ASpan}        & {58.4}$_{+5.42\%}$      & {74.1}$_{+3.49\%}$      & {84.8}$_{+2.05\%}$       \\ \midrule
\multicolumn{3}{l}{QuadT~\cite{quadtree}~\tiny{ICLR'22}}                   & 54.6       & 70.5       & 82.2       \\
\multicolumn{3}{l}{OETR+QuadT~\tiny{AAAI'22}}        & {55.4}$_{+1.47\%}$      & {71.2}$_{+0.99\%}$      & {82.1}$_{-0.01\%}$       \\
\rowcolor[rgb]{.9,.95,.98}\multicolumn{3}{l}{MESA\_QuadT}        & 57.9$_{+6.04\%}$       & 73.1$_{+3.69\%}$       & 83.9$_{+2.07\%}$       \\ \midrule
\multicolumn{3}{l}{LoFTR~\cite{loftr}~\tiny{CVPR'21}}                   & 52.8       & 69.2       & 81.2       \\
\multicolumn{3}{l}{OETR+LoFTR~\tiny{AAAI'22}}        & {54.8}$_{+3.79\%}$      & {70.3}$_{+1.59\%}$      & {81.8}$_{+0.73\%}$       \\
\rowcolor[rgb]{.9,.95,.98}\multicolumn{3}{l}{MESA\_LoFTR}        & 56.8$_{+7.58\%}$       & 72.2$_{+4.34\%}$       & 83.3$_{+2.59\%}$       \\ \midrule
\multicolumn{3}{l}{DKM~\cite{dkm}~\tiny{CVPR'23}}                   & {60.4}       & {74.9}       & \uline{85.1}       \\
\multicolumn{3}{l}{OETR+DKM~\tiny{AAAI'22}}        & \uline{60.6}$_{+0.33\%}$      & \uline{75.1}$_{+0.27\%}$      & {84.9}$_{-0.24\%}$       \\
\rowcolor[rgb]{.9,.95,.98}\multicolumn{3}{l}{MESA\_DKM}        & \textbf{61.1}$_{+1.16\%}$       & \textbf{75.6}$_{+0.93\%}$       & \textbf{85.6}$_{+0.59\%}$       \\
\bottomrule
\end{tabular}
}
\vspace{-0.5em}
\caption{\textbf{Relative pose estimation results (\%) on MegaDepth-1500 benchmark.} The \textbf{best} and \uline{second} results are highlighted.}\label{tab:PAUC_MD}
\vspace{-1.5em}
\end{table}

\boldparagraph{Results.} In \cref{tab:PAUC_MD}, MESA achieves impressive precision improvement for semi-dense baselines, up to $+7.58\%$ in AUC@$5^\circ$.
The improvement for the dense method is not as notable as that for semi-dense methods, because lots of repetitiveness in MegaDepth leads to hard area matching, decreasing the accuracy of DKM.
However, MESA\_DKM does set a new \textit{state-of-the-art} for outdoor pose estimation. 
MESA also achieves better results than OETR by finding more detailed area matches, whereas OETR solely estimates covisible areas.

\subsection{Visual Odometry}
We also evaluate our method on the visual odometry using the KITTI360~\cite{KITTI360} dataset, which densely estimates the camera motion in the self-driving scene. We select four sequences with few moving objects from the total dataset.

\boldparagraph{Experimental setup.} The parameter $T_{E_{max}}$ is set as $0.25$. The area image size is $480 \times 480$. Our method is combined with four baselines~\cite{aspanformer,quadtree,loftr,dkm} to demonstrate its effectiveness. All baselines are trained on ScanNet~\cite{scannet}. Following~\cite{S2LD}, we report the relative pose errors (RPE), including the rotational error ($R_{err}$) and translation error ($t_{err}$). 

\boldparagraph{Results} \cref{tab:VO} shows that our method achieves prominent precision improvement for DKM~\cite{dkm}, ASpan~\cite{aspanformer} and LoFTR~\cite{loftr} in all sequences.
For QuadT~\cite{quadtree}, its performances increases in most cases with the aid of MESA. Two cases of accuracy declines may caused by its quadtree attention mechanism~\cite{quadtree}, which evenly splits the area image leading to destruction of semantic integrity, especially in the driving scene with strong symmetry. Our method combined with DKM achieves the best performance, proving the benefits of MESA for the dense matching method.

\begin{table}[!t]
\resizebox{\linewidth}{17mm}{
\begin{tabular}{cllllllllll}
\toprule
\multicolumn{3}{c}{\multirow{2}{*}{Method}} & \multicolumn{2}{c}{Seq. 00} & \multicolumn{2}{c}{Seq. 02} & \multicolumn{2}{c}{Seq. 05} & \multicolumn{2}{c}{Seq. 06} \\ \cmidrule(r){4-5} \cmidrule(r){6-7} \cmidrule(r){8-9} \cmidrule(r){10-11}
\multicolumn{3}{c}{}                                  & $R_{err}$ $\downarrow$      & $t_{err}$ $\downarrow$       & $R_{err}$ $\downarrow$      & $t_{err}$ $\downarrow$       & $R_{err}$ $\downarrow$      & $t_{err}$ $\downarrow$      & $R_{err}$ $\downarrow$      & $t_{err}$ $\downarrow$      \\ \midrule
\multicolumn{3}{c}{DKM~\cite{dkm} }                              & \nd 0.046        & \nd 0.569        & \nd 0.102        & \nd 0.683        & \nd 0.048        & \nd 0.623        & \nd 0.050        & \nd 0.533        \\
\multicolumn{3}{c}{MESA\_DKM}                          & \fs0.039        & \fs 0.499        & \fs 0.051        & \fs 0.611        & \fs 0.041        & \fs 0.546        & \fs 0.044        & \fs 0.488        \\ \midrule
\multicolumn{3}{c}{ASpan~\cite{aspanformer}}                           & \nd 0.152        & \nd 2.412        & \nd 0.149        & \nd 2.358        & \nd 0.154        & \nd 2.523        & \nd 0.146        & \nd 1.887        \\
\multicolumn{3}{c}{MESA\_ASpan}                        & \fs 0.130        & \fs 2.079        & \fs 0.141        & \fs 2.327        & \fs 0.131        & \fs 2.227        & \fs 0.132        & \fs 1.671        \\ \midrule
\multicolumn{3}{c}{QuadT~\cite{quadtree}}                             & \nd 0.145        & \nd 2.307        & \nd 0.144        & \nd 2.322        & \nd 0.135        & \nd 2.269        & \nd 0.137        & \nd 1.762        \\
\multicolumn{3}{c}{MESA\_QuadT}                        & \fs 0.134        & \fs 2.305        & \fs 0.140        & \rd 2.420        & \fs 0.123        & \fs 2.131        & \fs 0.135        & \rd 1.915        \\ \midrule
\multicolumn{3}{c}{LoFTR~\cite{loftr}}                             & \nd 0.132        & \nd 2.023        & \nd 0.124        & \nd 1.913        & \nd 0.117        & \nd 1.930        & \nd 0.126        & \nd 1.572        \\
\multicolumn{3}{c}{MESA\_LoFTR}                        & \fs 0.116        & \fs 1.703        & \fs 0.114        & \fs 1.658        & \fs 0.109        & \fs 1.654        & \fs 0.120        & \fs 1.493        \\ \bottomrule
\end{tabular}
}
\vspace{-0.5em}
\caption{\textbf{Visual odometry results on KITTI360.} The \colorbox{colorSnd}{original} results of baselines and \colorbox{colorFst}{\bf better} or \colorbox{colorTrd}{worse} results achieved by our method are highlighted. All baselines are trained on ScanNet~\cite{scannet}.} \label{tab:VO}
\end{table}

\subsection{Ablation Study}
To evaluate the effectiveness of our design, we conduct a comprehensive ablation study for components of MESA.

\begin{table}[t]
\resizebox{\linewidth}{!}{
\begin{tabular}{ccccccc}
\toprule
\multicolumn{3}{l}{Method}               & AOR~$\uparrow$   & AMP@0.6~$\uparrow$ & PoseAUC@5~$\uparrow$ & AreaNum~$\uparrow$ \\ \midrule
\multicolumn{3}{l}{MESA\_ASpan (\textbf{Ours})}          & 72.75 & 89.09   & 27.50       & 3.47     \\ \hdashline \noalign{\vskip 1pt}
\multicolumn{3}{l}{ w/~\textit{CSD}}           & 69.23 & 84.21   & 26.78      & 2.73    \\
\multicolumn{3}{l}{ w/~\textit{DesSim.}~\cite{pats}}           & 63.71 & 62.91   & 26.05      & 2.65     \\
\multicolumn{3}{l}{ w/~\textit{SEEMSeg.}~\cite{SEEM}}          & 70.58 & 85.52   & 26.18      & 1.67     \\
\multicolumn{3}{l}{ w/~$\arg\min E_{self}$} & 70.98     & 87.56       & 26.96          & 2.90        \\ \bottomrule
\end{tabular}
}
\vspace{-0.5em}
\caption{\textbf{Ablation study.} Three variants of MESA\_ASpan are evaluated for area matching and pose estimation on the ScanNet1500 to demonstrate the importance of various components.} \label{tab:ASR}
\vspace{-1.2em}
\end{table}

\boldparagraph{Area Graph Construction.} To justify the AG of MESA, we adopt a naive approach to match areas, which is comparing area similarity densely (CSD). In particular, we first select areas with proper size from all SAM areas of two images. The similarity of each area to all areas in the other images is then calculated and area matches with the greatest similarity is obtained. The comparison results are summarised in Tab.~\ref{tab:ASR}. As AG can generate more proper areas for matching, MESA w/ CSD gets less area matches.
Thus, the area and point matching performance is also decreased by CSD. Moreover, CSD results in a significant increase in time of area matching ( $\sim \times 10$ slower than MESA), due to its inefficient dense comparison.

\boldparagraph{Area Similarity Calculation.}
In contrast to our classification formulation for area similarity calculation, another straightforward method~\cite{pats} involves calculating the distance between learning descriptors of areas.
Thus, we replace our learning similarity with descriptor similarity in~\cite{pats} (\textit{DesSim}) and conduct experiments in ScanNet to investigate the impact.  
The results are summarized in \cref{tab:ASR}, including the area number per image, area matching and pose estimation performance.
Overall, the performance of \textit{DesSim} experiences a noticeable decline, due to poor area matching precision, indicating the effectiveness and importance of proposed learning similarity calculation.

\boldparagraph{Image Segmentation Source.}
We relay on SAM to achieve areas with implicit semantic, whose outstanding segmentation precision and versatility contribute to our leading matching performance.
However, areas can also be obtained from other segmentation methods.
Therefore, to measure the impact of different segmentation 
sources, we exchange the segmentation input from SAM~\cite{sam} with that from SEEM~\cite{SEEM} (\textit{SEEMSeg.}) and evaluate the performances. In \cref{tab:ASR}, MESA with \textit{SEEMSeg.} gets a slight precision decline and fewer areas compared with SAM, leading to decreased pose estimation results. These results indicates that the advanced segmentation favors our methods. Notably, MESA with \textit{SEEMSeg.} also achieves slight improvement for ASpan, proving the effectiveness of MESA.

\begin{figure}[!t]
\centering
\includegraphics[width=\linewidth]{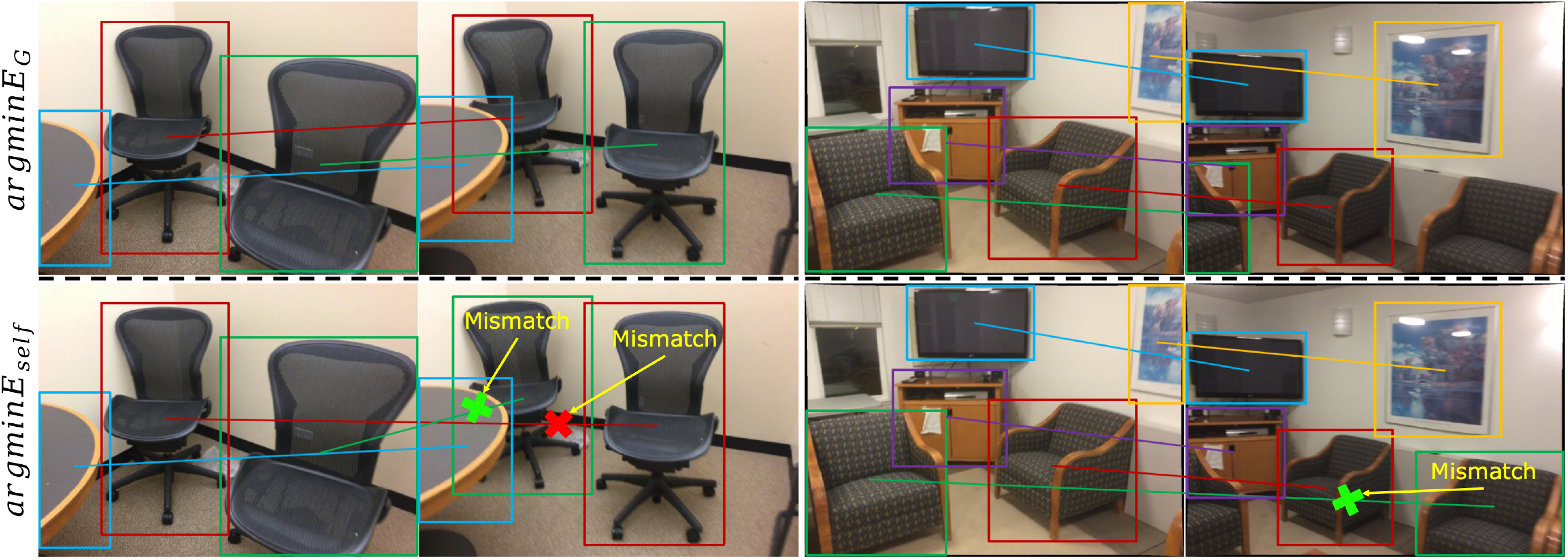}
\caption{\textbf{The qualitative comparison of Global Energy Refinement.} As AG structures of both images are considered by $E_G$, objects with the same apparent can be distinguished according to their neighbors, which are mismatched by $\arg\min E_{self}$, revealing the robustness of $\arg\min E_G$ under repetitive patterns.}
\vspace{-1.2em}
\label{fig:e_qs}
\end{figure}

\boldparagraph{Global Energy Refinement.}
After \textit{Graph Cut}, the proposed global matching energy for the final area matching refinement considers structures of both AGs of the input image pair. To show the importance of this dual-consideration, we replace the global energy with naive $E_{self}$ in \cref{eq:e_self} ($\arg\min E_{self}$) and evaluate the performance. In \cref{tab:ASR}, the refinement relying on $E_{self}$ produces decreased area matching precision and a subsequent decline in pose estimation performance, due to inaccurate area matches especially under repetitiveness. The qualitative results shown in \cref{fig:e_qs} further indicate the better robustness of global energy under repetitiveness due to dual graph structure capture.

\section{Conclusion}
In this paper, we propose MESA as an approach for accurate, robust, and practical area matching, aiming at effective feature matching redundancy reduction. 
To this end, MESA first utilizes the high-level image understanding capability of SAM, an advanced foundation model for image segmentation, to obtain informative image areas.
Then, we propose a novel Area Graph (AG) to model the spatial structure and scale hierarchy of image areas. By formulating the area matching on graphical models converted by AG, we adopt the \textit{Graph Cut} algorithm and a proposed learning area similarity model to establish area matches. We further consider the AG structures for both images to introduce the global matching energy refinement, which enables robust and exact area matching, promoting precise point matching.
In extensive experiments, MESA achieves impressive area matching accuracy and significantly improves feature matching performance for various point matching methods. 

\boldparagraph{Acknowledgement.} This work has been funded in part by the NSFC grants 62176156, and SJTU Trans-med Awards Research 20220101.

{
    \small
    \bibliographystyle{ieeenat_fullname}
    \bibliography{main}
}

\clearpage
\setcounter{page}{1}
\maketitlesupplementary

In the following, we provide additional details about the proposed area matching method, MESA. \cref{sec:ID} describes specific implementation details of our method. \cref{sec:cca} analyzes the computation complexities of main components in MESA. \cref{sec:aas} gives more insights and ablation studies, that motivates our design.  \cref{sec:lfw} states the limitation of MESA and our future work.

\section{Implementation Details}\label{sec:ID}
In this section, we provide sufficient details about the implementation of MESA for reproduction, including detailed operations in Graph Completion (\cref{sec:gcd}), training details about our Learning Area Similarity (\cref{sec:lasd}) and complete parameter settings in Graphical Area Matching (\cref{sec:gamd}).

\subsection{Graph Completion}\label{sec:gcd}
The detailed algorithm for our graph completion is depicted in \cref{alg:gc}, which takes initial AG ($\mathcal{G}_{ini}$) as input and outputs the final AG ($\mathcal{G}$) with scale hierarchy. Additionally, we describe the area cluster and two main area operations in the algorithm as follows.

\boldparagraph{Area Cluster.}
For orphan nodes in each level, we cluster them based on their area centers to decide which operation will be performed on them.
We use the k-means algorithm with elbow method~\cite{kmeans} to determine the cluster number. The candidate cluster number is set as $\{1,\dots,n\}$, where $n$ is the number of orphan nodes in the current level. This algorithm is fed with area centers and outputs labeled ones.

\begin{figure}[!t]
\centering
\includegraphics[width=\linewidth]{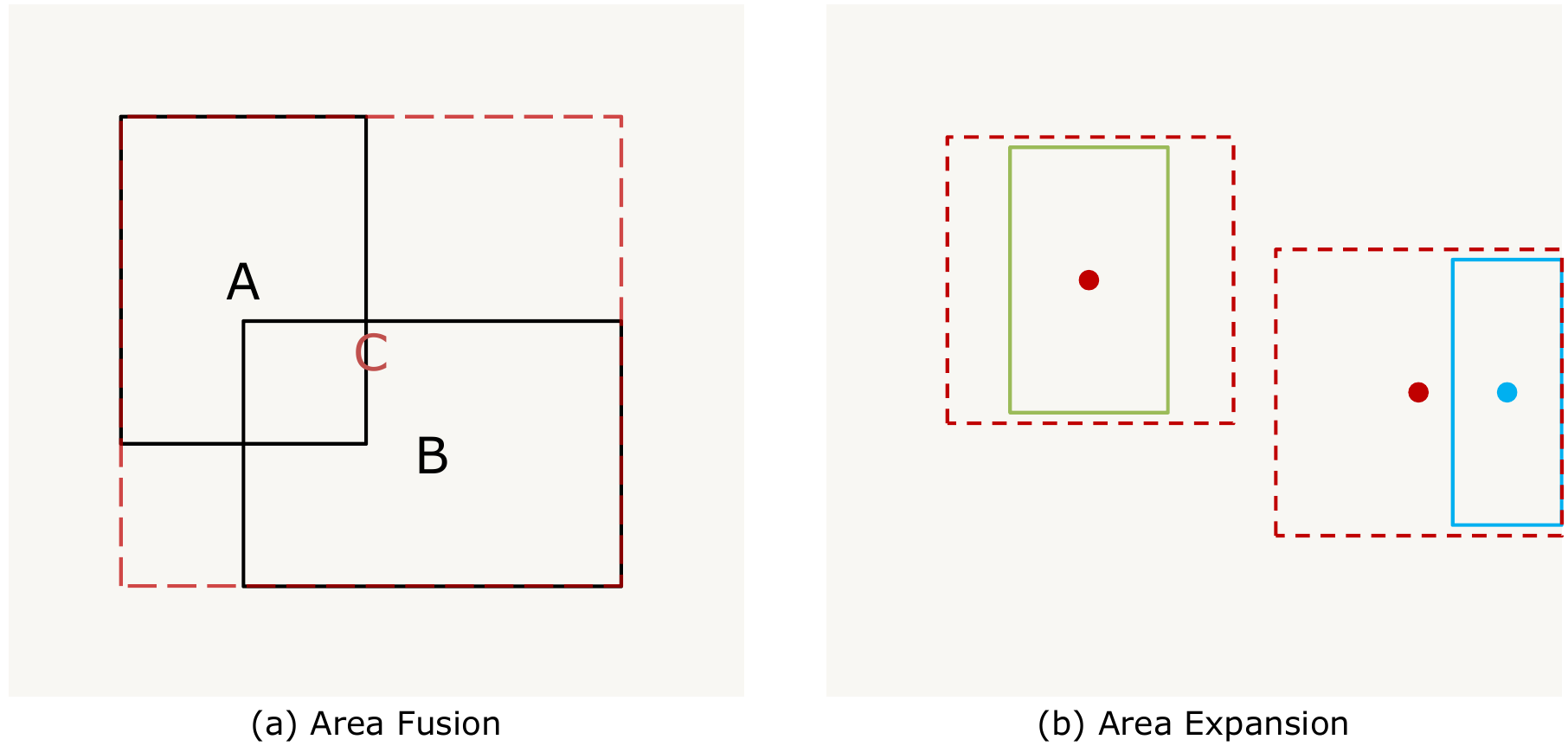}
\caption{\textbf{The Area Fusion and Area Expansion.} \textbf{(a)} Area fusion is to achieve the smallest area (\textit{C}) containing the input areas (\textit{A} and \textit{B}). \textbf{(b)} Generally, area expansion is to fix the \textcolor[rgb]{.61,.73,.35}{original area} center and expand its size to the smallest size of the next level. When the \textcolor[rgb]{.0,.69,.94}{original area} is too close to the image boundary, we will move the area center to keep the expanded area inside the image.}
\vspace{-1.2em}
\label{fig:ao}
\end{figure}

\boldparagraph{Area Fusion and Expansion.}
Area fusion and expansion are key operations in our graph completion algorithm.
Specifically, area fusion is to find the largest outer rectangle of the two areas as the new area, as depicted in \cref{fig:ao} (a). Due to the careful threshold settings of our area level, the fused area size will exceed current level and be awaited for subsequent operations.
On the other hand, the expansion operation is to expand the area to the next level size (\cref{fig:ao} (b)). In particular, suppose the lower bound of size for the next level is $s^2$, if both of the area width and height are smaller than $s$, we expand the height and width of the area to $s$, keeping the area center fixed. Otherwise if area width $w \geq s$, we let the area height $h = s^2/w$, keeping the area center fixed, and vice versa.
The above operations are performed when the expanded area is inside the image. On the other hand, if the expanded area is outside the image, the area center will be moved as shown in \cref{fig:ao} (b).

\subsection{Learning Area Similarity.}\label{sec:lasd}
We propose the learning area similarity model for area similarity calculation, which is the basis of our graphical area matching. In this section, we describe the training protocol of this model, including supervision and training details.

\boldparagraph{Supervision.}
We generate regular area images from both indoor and outdoor datasets~\cite{scannet,megadepth} as training data through the proposed method in \cref{sec:ag}.
Then the area pairs with more than $30\%$ overlap are collected.
For each image patch $p_i$ in these pairs, its ground truth activity $\sigma_{i}^{gt}$ is set as $1$, if more than $60\%$ pixels in it have correspondences in the other area image, and $0$ otherwise.
As our classification formulation, we use the binary cross entropy ($BCE$) of each patch classification to form the loss function of area similarity calculation ($L_{asc}$).
\begin{equation}
    L_{asc} = \frac{1}{Z}\sum_{i}^{Z}BCE(\sigma_{i}^{gt}, \sigma_{i})
\end{equation}
Based on this loss, our network can learn to achieve the similarity between two area images.

\begin{algorithm2e}[t]
    \normalem
    \caption{Graph Completion}\label{alg:gc}
    \KwIn{$\mathcal{G}_{ini}=\langle \mathcal{V}_{ini}, \mathcal{E}_{ini}\rangle$}
    \KwOut{$\mathcal{G} = \langle \mathcal{V}, \mathcal{E}\rangle$}

    \For{$l$ in $[0, L-1]$}
    {
        initial orphan node set $\mathcal{O} = \varnothing$\;
        \For{$v_i \in \{v_i|l_{a_i} = l\}$}
        {
        \uIf{$v_i$ has no parent}
        {
            add $v_i$ into $\mathcal{O}$\;
        }
        }
        cluster the nodes in $\mathcal{O}$ based on their area centers\;
        \For{each node cluster $\mathcal{C}_h=\{v_k\}_{k=0}^C$}
        {
            \eIf{$C \ge 2$}
            {
                \For{each $v_k \in \mathcal{C}_h$}
                {
                    \uIf{$v_k$ has not been fused}
                    {
                        fuse area $a_k$ with its nearest neighbor $a^n | v^n \in \mathcal{C}_h$:
                        $a^f = F({a}_k, a^n)$\;
                        generate higher level node $v^f$ for $a^f$\;
                        add $v^f$ into $\mathcal{V}_{ini}$\;
                        form edges by Link Prediction: $\{e_h\}_h = LP(v^f, \mathcal{V}_{ini})$\;
                        add $\{e_h\}_h$ into $\mathcal{E}_{ini}$\;
                    }
                }
            }
            {
            Update the single node $v_0$: $v^u_0 = Up(v_0)$\;
            construct edges: $\{e_j\}_j = LP(v^u_0, \mathcal{V}_{ini})$\;
            add $\{e_j\}_j$ into $\mathcal{E}_{ini}$\;
            }
        }
    }
    $\mathcal{E} = \mathcal{E}_{ini}$\ \;
    $\mathcal{V} = \mathcal{V}_{ini}$\ \;
    output the updated AG: $\mathcal{G} = \langle \mathcal{V}, \mathcal{E}\rangle$\;
\end{algorithm2e}

\boldparagraph{Training Details.}
Following previous point matching work~\cite{loftr}, the proposed learning similarity module is trained utilizing MegaDepth~\cite{megadepth} and ScanNet~\cite{scannet} dataset.
For each image pair sampled in the datasets in every epoch following \cite{loftr}, we collect up to 5 area image pairs for training.
Fortunately, there is no need to train this model from scratch, thanks to the similar objective between the coarse point matching and the area similarity calculation, both of which aims at patch-level similarities.
Therefore, we adopt the pretrained weights of coarse level feature operation in ASpan~\cite{aspanformer} on our network.
With the modified output head, our network is fine-tuned using AdamW~\cite{adamw} on 2 NVIDIA RTX 4090 GPUs.
We use the same learning rate and batch size settings as ASpan and LoFTR~\cite{loftr} during training. This model can converge within 3 epochs.

\subsection{Graphical Area Matching}\label{sec:gamd}
In this section, we describe some implementation details of our graphical area matching. First, we illustrate the complete forward process of area matching. 
Then, we specific the parameter settings we recommend and utilized in our experiments. Next, the details about area image cropping is presented, which is cropping area images from original images to serve as input images for the combined point matcher. The post-processing of point matching within areas is also illustrated. 

\boldparagraph{Area Matching Process.}
Given the image pair ($I_0, I_1$) and their area graphs ($\mathcal{G}^0, \mathcal{G}^1$), we first collect area nodes with specific size level $l_{a^*}$ from $\mathcal{G}^0$ as the \textit{source nodes}, which have proper sizes for subsequent point matching. Then, area matches are found for these nodes from $\mathcal{G}^1$ by our method. Afterwards, we exchange the two images to repeat the above operations. The final area matches are the common matching results.

\boldparagraph{Parameter Settings.}
We describe the common parameters for different scenes here.
During the AG construction, the input image are resized to $640 \times 480$.
The aspect ratio threshold $T_r = 4$ and minimal size threshold is $T_s = 80^2$.
The number of area size threshold is $4$ and specific $TL_i$s are $\mathsf{80^2, 130^2, 256^2, 390^2, 560^2}$.
The $\delta_l$ is $0.1$ and $\delta_h$ is $0.8$.
In graphical area matching, the $\lambda$ in \cref{eq:e} is $0.1$.
The area similarity threshold $T_{as} = 0.05$.
The energy balance weights ($\mu, \alpha, \beta, \gamma$) in \cref{eq:eg} are $\mathsf{4,2,2,2}$.  The specific area level $l_{a^*}$ for point matching is $1$. The $T_{Er}$ in \cref{eq:collectarea} is $0.1$.
Other parameters are specified for different scenes as described in our paper.

\boldparagraph{Area Image Cropping and Point Matching.}
Following the A2PM framework~\cite{sgam}, the next step after area matching is to perform point matching inside area matches. Thus, these area matches need to be achieved through image cropping. At the same time, one of the benefits of A2PM is the input image with high resolution that can be provided to the point matcher. Therefore, the cropping is performed on original images with the highest resolution. A straightforward cropping approach is to crop areas along the exact bounding box and then resize them to default input size of point matchers.
However, as area sizes can be quite different from default input sizes of point matchers, direct cropping and resize introduce severe distortion for point matching, resulting in decreased matching precision (cf. \cref{sec:abas}). In order to address this issue, we propose to crop area image by considering the aspect ratio. To be specific, we force the cropped area image to possess the same aspect ratio as the input size of the point matcher, while trying to keep the area center unchanged. If the area respect ratio ($W_a/H_a$) is larger than the input aspect ratio ($W_i/H_i$), we fix the width $W_A$ of area image and expand the height $H_a$ to $(W_a\times H_i)/W_i$. Otherwise, we fix the height and expand the width to $(H_a\times W_i)/H_i$. Moreover, as solid feature points often cluster in the boundaries of objects, we set a spread ratio ($r_s=1.2$) to slightly increase the cropping size, allowing for more precise correspondences on the boundaries. Finally, if the cropped area exceeds the original image, we will move its center to keep it inside the image, similar to our Area Expansion in \cref{fig:ao}. 

For point matching inside area matches, in practice, we empirically set the input size of point matchers to a square, \ie the input aspect ratio is $1$, as it can lead to statistically the smallest area size adjustment. However, as the learning models are sensitive to training size and scale (the area input has different scale with original image), especially the Transformer~\cite{patch}, the performance of baselines are decreased, as we shown in Sec.~\ref{sec:abas}. 
Therefore, we {fine-tune} the baselines under the square sizes with ground truth area matches generated by our method, \eg $640\times 640$ area matches in indoor scenes, to achieve similar accuracy with original models, before combined with MESA. 

\boldparagraph{Post-processing.}
In SGAM~\cite{sgam}, \textit{Global Match Collection} (GMC) is proposed to collect precise point matches through globally entire-image matching, which significantly improve the matching precision~\cite{sgam}. Therefore, we adopt this module in our method as well and set the occupancy ratio as $0.6$. However, as SAM enables areas throughout the image, MESA usually achieves enough area matches to cover the whole overlap between images. Thus, this module is only activated in few cases, but also helpful to precise matching. As we mentioned before, we also adopt GAM~\cite{sgam} as the post-processing of area matching.

\section{Computation Complexity}\label{sec:cca}
Here, we analyze the computation complexity of proposed graphical area matching.
\subsection{Area Similarity Calculation}
Firstly, area similarity calculation is performed to achieve the required node energies in the graph, serving as the prerequisite of our graphical area matching.
Suppose we have two AGs, $\mathcal{G}^0 $ and $ \mathcal{G}^1$, for the input image pair, $\mathcal{G}^0$ gets $N$ nodes ($|\mathcal{V}^0|=N$) and $\mathcal{G}^1$ gets $M$ nodes ($|\mathcal{V}^1|=M$). Therefore, the dense graph energy calculation needs $M \times N$ times similarity calculation. However, owing to the similarity conditional independence of ABN (Sec.~\ref{sec:abn}), the actual number ($M' \times N'$) of similarity calculation is smaller than $M \times N$, as $N' < N$. Nevertheless, directly setting children pair similarities as $0$ is too rough (\cref{eq:abn}), as large scale differences also leads to near-zero similarity between areas. In practise, we only set the related similarities of \textit{next level children} as $0$ for area matching accuracy and the efficiency from ABN is still helpful to our approach.  Moreover, we only care about the similarities between source nodes in $\mathcal{G}^0$ and other nodes in $\mathcal{G}^1$, because we collect source nodes with specific level from $\mathcal{G}^0$ to match, \eg, usually $3\sim 4$ areas in indoor scene and less in outdoor scene. Therefore, we have $M' < M$. Similarly, in the case of duality, \ie, collecting source nodes from $\mathcal{G}^1$ to match, we only need to perform a few supplementary calculations, as similarities are symmetric and reusable.
Thus, the real computation complexity of area similarity computation is $O(M' \times N')$, where $M' \times N' < M \times N$.

\subsection{Edge Energy Calculation}
Except the node energy calculation, the edge energy is also needed to be determined for \textit{Graph Cut}. The computation complexity of edge energy calculation is related to edge number of $\mathcal{G}^0$ and $\mathcal{G}^1$. Assume $|\mathcal{E}_0|=E$ and $|\mathcal{E}_1|=K$, the specific computation complexity is $O(E+K)$. 

\subsection{Global Energy Minimization}
In our global energy minimization for area matching refinement, the matching energy of parent, children and neighbour pairs all need to be calculated. Taking parent matching energy for example, we derive its computation complexity as follows. Suppose $n$ nodes are achieved as match candidates through \textit{Graph Cut} and each node gets $Q_i,\;i\in(0,n]$ parent nodes, there are $Q_i \times V$ node similarities need to be accessed (as the similarity calculation is finished), where $V$ is the parent node number of the source node. Hence, the total computation complexity for parent matching energy in global energy minimization is $O(\sum_i^n Q_i \times V)$. The children matching energy and neighbour matching energy are similar.
As $n$ is the number of node after \textit{Graph Cut}, it is small in most cases, \eg, usually $<3$ area nodes. Moreover, the number of parent nodes (or children, neighbour nodes) is also limited. Therefore, the computation complexity for global energy minimization is acceptable in practise.

\begin{table*}[!t]
\centering
\resizebox{\linewidth}{15mm}{
\begin{tabular}{cclllclll}
\toprule
Input Image Size          & {Semi-Dense Matcher} & {AUC@5~$\uparrow$} & {AUC@10~$\uparrow$} & \multicolumn{1}{c}{AUC@20~$\uparrow$} & \multicolumn{1}{c}{Dense Matcher} & \multicolumn{1}{c}{AUC@5~$\uparrow$} & \multicolumn{1}{c}{AUC@10~$\uparrow$} & \multicolumn{1}{c}{AUC@20~$\uparrow$} \\ \cmidrule(r){1-1} \cmidrule(r){2-2} \cmidrule(r){3-5} \cmidrule(r){6-6} \cmidrule(r){7-9}
\multirow{2}{*}{$640\times640$} & ASpan                                  & 26.33                     & 45.98                     & 62.12                       & DKM           & 28.63                     & 50.84                      & 68.97                     \\
                         & \ours MESA\_ASpan                            & \ours 27.51$_{+4.48\%}$                    & \ours 47.47$_{+3.24\%}$                      & \ours 65.04$_{+4.70\%}$                       & \ours MESA\_DKM     & \ours 33.42$_{+16.73\%}$                     & \ours \ours 55.04$_{+8.26\%}$                      & \ours 71.98$_{+4.36\%}$                      \\ \hdashline \noalign{\vskip 1pt}
\multirow{2}{*}{$480\times480$} & ASpan                                  & 21.38                     & 40.53                      & 58.74                       & DKM           & 28.85                     & 50.06                      & 68.20                       \\
                         & \ours MESA\_ASpan                            & \ours 24.01$_{+12.30\%}$                     & \ours 43.32$_{+6.88\%}$                      & \ours 60.99$_{+3.83\%}$                       & \ours MESA\_DKM     & \ours 33.00$_{+14.38\%}$                     & \ours 54.04$_{+7.95\%}$                      & \ours 71.02$_{+4.13\%}$                      \\ \hdashline \noalign{\vskip 1pt}
\multirow{2}{*}{$320\times320$} & ASpan                                  & 8.95                      & 21.18                      & 37.68                       & DKM           & 27.55                     & 48.20                      & 65.96                      \\
                         & \ours MESA\_ASpan                            & \ours 9.60$_{+7.26\%}$                       & \ours 22.17$_{+4.67\%}$                      & \ours 38.70$_{+2.71\%}$                        & \ours MESA\_DKM     & \ours 31.43$_{+14.08\%}$                     & \ours 52.56$_{+9.05\%}$                      & \ours 69.99$_{+6.11\%}$                      \\ \bottomrule
\end{tabular}
}
\caption{\textbf{Ablation study of area image size.} We investigate the performance impact of three different input image sizes for both semi-dense and dense methods, along with our MESA combined with them. For point matchers, the input image size is the size of resized original image. For our method, the input image size is the area image size. The pose estimation AUC$@5^{\circ}/10^{\circ}/20^{\circ}$ are reported for evaluation.}\label{tab:ABAS}
\end{table*}

\begin{table}[!t]
\resizebox{\linewidth}{10mm}{
\begin{tabular}{ccccc}
\toprule
Method                       & Cropping Approach     & AUC@5~$\uparrow$ & AUC@10~$\uparrow$ & AUC@20~$\uparrow$ \\ \midrule
\multirow{2}{*}{MESA\_ASpan} & \textit{OAR}             &  24.67 & 43.72  & 61.29  \\
                             & \textit{ARPM} & \bf 27.51 & \bf 47.47  & \bf 65.04  \\ \hdashline \noalign{\vskip 1pt}
\multirow{2}{*}{MESA\_DKM}   & \textit{OAR}             & 30.19 & 51.49  & 68.79  \\
                             & \textit{ARPM} & \bf 33.42 & \bf 55.04  & \bf 71.98 
                             \\ \bottomrule
\end{tabular}
}
\caption{\textbf{Ablation study of area image cropping.} Two different image cropping methods are compared for the proposed MESA. Both semi-dense and dense point matchers are combined for evaluation. We report the pose estimation AUC$@5^{\circ}/10^{\circ}/20^{\circ}$ and the \textbf{best} results of two series are highlighted respectively.}\label{tab:ABAC}
\end{table}

\section{Additional Ablation Study}\label{sec:aas}
In this section, we examine the performance impact of more components in MESA, including the input image size (\cref{sec:abas}), image cropping approach (\cref{sec:abic}) and energy parameter setting (\cref{sec:abep}).

\subsection{Ablation Study on Input Image Size}\label{sec:abas}
Input image size is a sensitive parameter for feature matching, as the larger the image size, the higher the resolution and the richer the information in the image. At the same time, especially for transformer-based point matchers~\cite{aspanformer,casmtr,loftr}, different input image sizes produce widely varying matching results. To investigate the effectiveness of our MESA under different image sizes, we construct experiments on ScanNet1500 benchmark~\cite{scannet}. In particular, we combine our MESA with both semi-dense point matcher ASpan~\cite{aspanformer} and dense point matcher DKM~\cite{dkm} to estimate relative pose from input images with three different sizes: $640 \times 640$, $480 \times 480$ and $320 \times 320$. It is noteworthy that we choose square sizes, because areas have different raw sizes and square input leads to smallest distortion from resize.
The original point matchers (ASpan and DKM) are trained in $640\times 480$ and fine-tuned in $640 \times 640$, using ground truth area matches generated by our method.
We compare MESA\_baselines with original baselines under the same input size to demonstrate the effectiveness of our methods. The results are summarised in Tab.~\ref{tab:ABAS}. The fine-tuned point matchers achieve comparable results in $640 \times 640$ with original ones in $640\times 480$, proving the fine-tuning is an effective way to remove the impact from inconsistent image sizes and scales between training and testing.  Overall, MESA effectively increases the performance for both point matchers under all input sizes.
For the semi-dense point matcher ASpan, the accuracy decreases significantly as the input size decreases, whereas the improvement achieved by MESA remains noticeable. Moreover, the improvement under $480\times 480$ is much better than under $640\times 640$ ($12.30\% \;vs.\; 4.48\%$), revealing the benefits of high resolution provided by MESA.  
On the other hand, DKM is much more robust under different input sizes and only gets slight performance declines with smaller input sizes. Meanwhile, our MESA achieves impressive improvements under all input sizes proving the effectiveness of MESA. Furthermore, it is worth noting that MESA\_DKM achieves better results with smaller input size than original DKM \eg, $33.00$ for MESA\_DKM under $480\times 480$ is better than $28.63$ for DKM under $640\times640$, indicating the superiority of MESA.
In sum, MESA enables effective matching redundancy reduction which allows for high resolution input with less matching noises, leading to advanced feature matching under different input image sizes.

\subsection{Ablation Study on Image Cropping}\label{sec:abic}
The image cropping is a trivial yet important operation for the A2PM framework, as different cropping approaches lead to different image resolutions and distortions. Here, we construct experiments to investigate the impact of different cropping approaches. To be specific, we compare two different cropping methods descried in \cref{sec:gamd}: \textbf{1)} the direct cropping method (\textit{OAR}), which crops with Original Aspect Ratios of areas; \textbf{2)} the cropping method with the Aspect Ratio of Point Matcher (\textit{ARPM}), which first expands the area to correspond with the aspect ratio of the point matcher input and then crop areas. The experiment is conducted on ScanNet1500~\cite{scannet} benchmark. We combine MESA with both semi-dense (ASpan) and dense (DKM) point matchers for complete comparison. Results are summarized in Tab.~\ref{tab:ABAC}. As we can seen that the ARPM cropping approach outperforms the OAR approach with a large margin for both MESA\_ASpan and MESA\_DKM, proving its superiority due to high resolution and less distortion. Therefore, we adopt the ARPM approach for area image cropping in MESA.

\begin{table}[!t]
\centering
\resizebox{\linewidth}{17mm}{
\begin{tabular}{cccccc}
\toprule
$E_{G}$ Parameters & $T_{E_{max}}$ & AOR~$\uparrow$   & AMP@$0.6$~$\uparrow$ & Pose AUC@$5^\circ$~$\uparrow$ & AreaNum~$\uparrow$ \\ \midrule
\multirow{3}{*}{\makecell[c]{$\mu=5,\;\alpha=2,$\\$\beta=2,\;\gamma=1$}}
& 0.35          & 61.76 &  65.54     &  \rd 23.57              & \nd 4.69    \\              
& 0.25          & 63.91 &  71.13     & 22.41              & 3.47    \\
& 0.15          & 60.44 & 62.57     & 21.46              & 3.27    \\ \hdashline \noalign{\vskip 1pt}
\multirow{3}{*}{\makecell[c]{$\mu=4,\;\alpha=2,$\\$\beta=2,\;\gamma=2$}}                     
& 0.35          & \fs 67.98 & \fs 80.09     & \nd 23.74              & \fs 5.76    \\
& 0.25          & \rd 64.94 & \rd 72.24     & \fs 24.01              & \rd 4.62    \\
& 0.15          & 61.74 & 65.50     & 23.55              & 3.86    \\ \hdashline \noalign{\vskip 1pt}
\multirow{3}{*}{\makecell[c]{$\mu=7,\;\alpha=1,$\\$\beta=1,\;\gamma=1$}}                     
& 0.35          & \nd 65.98 & \nd 78.10     & 22.71             & 3.27    \\
& 0.25          & 62.32 & 66.54     & 23.56              & 2.92    \\ 
& 0.15          & 60.32 & 64.38     & 22.37                 & 2.77      \\ \bottomrule
\end{tabular}
}
\caption{\textbf{Ablation study of global energy parameters.} We compare different parameter settings for global energy refinement in MESA\_ASpan and report the area matching performance, area number per image (AreaNum), and the pose estimation performance. Results are highlighted as \colorbox{colorFst}{\bf first}, \colorbox{colorSnd}{second} and \colorbox{colorTrd}{third}.}\label{tab:ABEP}
\end{table}

\begin{table}[!t]
\resizebox{\linewidth}{4mm}{
\begin{tabular}{cccccc}
\toprule
{Method} & LoFTR~\cite{loftr}   &  PATS~\cite{pats} & DKM~\cite{dkm}  & COTR~\cite{cotr} & MESA (Ours)  \\
Time(ms)/img  & 181.5 &  450.4 & 1318.5  & 18975.3 & 3092.4
                             \\ \bottomrule
\end{tabular}
}
\caption{\textbf{Time consumption comparison.} All methods run on $640\times 480$ images. }\label{tab:time}
\vspace{-1.2em}
\end{table}

\subsection{Ablation Study on Global Energy Parameters}\label{sec:abep}
The parameters for our global energy refinement mainly consists of global energy balance parameters ($E_G$ Parameters) in \cref{eq:eg} and the threshold parameter $T_{E_{max}}$.
The four $E_G$ Parameters reflect the importance of four energy terms, \ie, self matching energy, parent, children and neighbour matching energy.
The $T_{E_{max}}$ controls the maximum energy of the final match, the smaller it is the stricter the refinement.
Here, we construct experiments on ScanNet1500 to investigate the performance impact of these parameters. In particular, we compare three groups of $E_G$ Parameters and three groups of $T_{E_{max}}$ to evaluate their impact on MESA\_ASpan. The input size of ASpan is $480\times 480$. The area matching performance, pose estimation performance and area number per image are summarised in \cref{tab:ABEP}.
Generally, if two areas are matched, their parent, children and neighbour nodes should have high similarities due to spatial relationships between them.
At the same time, the self matching energy should still be an important reference in matching refinement.
Thus we choose three parameter settings including different weights on three kinds of node matching energies and different emphasis on self-matching energy.
The experiment results in Tab.~\ref{tab:ABEP} show that the weights of three parameter settings set to the same is better for area matching performance ($\alpha\!=\!\beta\!=\!\gamma~vs.~\alpha\!=\!\beta\!\not=\!\gamma$).
Giving sufficient consideration on global matching leads to accurate area matching along with best point matching performance ($\mu\!=\!4~vs.~\mu\!=\!7$).
Despite the semi-dense matcher is not sensitive to area matching accuracy, better area matching leads to higher pose estimation precision.
Therefore, we choose $[\mathsf{4,2,2,2}]$ as our energy setting.
On the other, the $T_{E_{max}}$ is a critical parameter as well. 
The smaller $T_{E_{max}}$ means stricter global matching energy request, but it may also mistake some accurate area matches when too small.
Different $E_G$ Parameter settings prefer different values of $T_{E_{max}}$ and $0.35$ suits the best for ours.


\section{Limitation and Future Work}\label{sec:lfw}
One limitation of MESA is its under-utilisation of SAM features. As we mentioned before, SAM possesses the high-level image understanding across a wide range of domains due to the massive training dataset and carefully designed models. Therefore, its image embedding is a extremely strong high-level representation, which has the potential to replace our learning similarity model. Then, the computation cost can be reduced as well. However, the naive attempt to use SAM features as descriptors of areas failed, possibly because the SAM segmentation pays more attention on intra-image contexts rather than inter-image ones like feature matching. Hence, the SAM feature needs further distillation for area matching, which will be a objective of our future work. 

On the other hand, as MESA fuses image areas based on their 2D distances, which is not equivalent to the 3D situations. Thus, some inconsistent area fusions between two images arise and lead to inaccurate point matching, \eg, shown in \cref{fig:fc}. Although the post-processing like GAM~\cite{sgam} may help, it also introduces extra computation cost. To address this issue, feature-guided fusion can be adopted, where the SAM feature can be employed and lead to consistent area fusion. 

Another limitation of MESA is the speed, which takes around $3$s per image for area matching as shown in \cref{tab:time}, limiting its performance in latency-sensitive applications like SLAM. 
In \cref{tab:time}, we also compare MESA with recent point matching methods with regard to running speed, all of them get $640 \times 480$ image as input and are implemented on a single NVIDIA 4090 GPU. Although MESA is slower than most point matchers, it is still faster than COTR~\cite{cotr}. As area matching is a pre-task for point matching, similar or even faster speed is the object of practical area matching for real-time tasks. It is also worthy to note that the speed of MESA is independent to image size (fixed area size is adopted in MESA), while these point matching methods will get significant increase in elapsed time when the image size is increased.
This drawback is mainly caused by multiple similarity calculations in MESA, where the utilisation of SAM features may be helpful as described above. At the same time, engineering technologies, like parallel area similarity computing, can facilitate our area matching.  We will investigate these possibilities in our future work.

\end{document}